\documentclass{article}
\usepackage{spconf, amsmath, graphicx, bbold, mathtools, amssymb, hhline, multirow, subfig, relsize}

\DeclarePairedDelimiter{\ceil}{\lceil}{\rceil}

\newcommand{\BS}[2]{\mathbb{BS}\left(#1,#2 \right)}
\newcommand{\subtag}[1]{\tag{\theparentequation#1}}

\title{Neural Network Training with Approximate Logarithmic Computations}
\name{Arnab Sanyal, Peter A.~Beerel, and Keith M.~Chugg\thanks{This work is supported in part by the National Science Foundation (CCF-1763747).}}
\address{Ming Hsieh Department of Electrical and Computer Engineering\\
University of Southern California\\
Los Angeles, California 90089, USA\\
\{arnabsan, pabeerel, chugg\}@usc.edu}

\begin{document}
%
\maketitle
\begin{abstract}
The high computational complexity associated with training deep neural networks limits online and real-time training on edge devices. 
This paper proposed an end-to-end training and inference scheme that eliminates multiplications by approximate operations in the log-domain which has the potential to significantly reduce implementation complexity.
We implement the entire training procedure in the log-domain, with fixed-point data representations. This training procedure is inspired by hardware-friendly approximations of log-domain addition which are based on look-up tables and bit-shifts. 
We show that our $16$-bit log-based training can achieve classification accuracy within approximately $1\%$ of the equivalent floating-point 
baselines for a number of commonly used datasets.
\end{abstract}
\begin{keywords}
Logarithmic Number System, Deep Neural Networks, Approximate Computation
\end{keywords}
\section{Introduction}
\label{sec:intro}

In recent years neural networks with hidden layers, or \textit{deep neural networks} (DNNs), have found widespread application in a large number of pattern recognition problems, notably speech recognition and computer vision \cite{deep_learning_book}. This resurgence in interest and application of neural networks has been driven by the availability of large data sets and increased computation resources. In particular, \textit{graphic processor units} (GPUs) provide a large number of hardware \textit{multiply-accumulate} (MAC) accelerators and are therefore widely used in the MAC-intensive process of training DNNs.  

Despite these advances, custom hardware accelerators have the potential to further improve the speed and energy efficiency in DNN training and inference modes. In a custom implementation one can co-design the computational circuitry, the simple control circuitry, and the memory architecture to achieve near full utilization of the hardware. This provides a potential advantage over GPUs which are more general purpose and may not efficiently utilize all the associated MAC hardware units during DNN processing. Most researchers have focused on accelerating inference processing with an eye towards using trained DNNs on edge computing devices (cf., \cite{ma2019pconv}) with a few more recent investigations \cite{sourya_jetcas, NullaNet} considering accelerated training. Hardware acceleration of training has the potential to reduce energy consumption in data centers and to provide learning directly on edge devices.  

There are a wide class of DNN complexity reduction methods based 
on sparsity, pruning, and quantization (cf., \cite{icann_fanout, kundu2019predefined, iclr2017_incQuant}). Complimentary to these approaches are methods of reducing the complexity of MAC units.  
Motivated by the fact that multiplier circuitry dominates the complexity of MAC units, several researchers have studied the use of \textit{logarithmic number system} (LNS) \cite{lns_three_times, sign_lns} wherein multiplications are replaced with additions. LNS methods have been proposed in communications \cite{map_lns}, processor design \cite{rom_less_lns}, re-configurable architectures \cite{fpga_lns}, and a number of signal processing applications \cite{lns_dsp}. The primary challenge of LNS-based MAC processing is the log-domain addition operation which has generally been addressed with functional approximation or \textit{look-up tables} LUTs.  
%
Preliminary work \cite{lns_old_hdware,oldest_lns_backprop} proposed an $8$ bit input, $16$ bit output LNS-based MAC using LUTs and claimed a $3.2\times$ improvement in area-delay product compared to an equivalent linear-domain MAC. This was investigated in the context of back-propagation, but was prior to resurgence of neural networks and therefore was not investigated in the context of modern datasets, larger networks, and other modern deep learning methods and conventions. 
Previous work also studied encoding weights
in LNS for inference \cite{stanford_lns} and proposed extensions to LNS MACs restricted to positive numbers \cite{stanford_arxiv}.
A more recent paper \cite{fair_log} implemented log-encoding on posits \cite{posit} but relied on conversions to and from the linear domain to perform addition. 

In this paper we propose end-to-end log-based training and inference of DNNs using approximate LNS computations. We generalize the bit-shift approximation in \cite{stanford_arxiv} to handle signed arithmetic and show that these bit-shift approximations are special cases of a LUT. To evaluate this approach we train a number of networks using fixed-point data representations in both the conventional linear domain and using our proposed approximate LNS computations based on both LUT and bit-shift approximations. Our results show that with 16-bit words and a 20-element LUT the degradation in training accuracy, relative to conventional floating point linear processing, is small (i.e., $\lesssim 1\%$ loss in accuracy).

The remainder of this paper is organized as follows - A brief description of LNS is provided in Section \ref{sec:lns}. Several approximations for LNS operations are developed in Section \ref{sec:appr}. Section \ref{sec:lognn} contains a description of the end-to-end training scheme of a neural network in log-domain as well as analysis relating bit-widths for fixed-point processing in the linear and log domains. Experimental results are summarized in Section \ref{sec:results} and conclusions provided in Section \ref{sec:conc}. 

\section{Logarithmic Number System}
\label{sec:lns}


In a LNS, a real number $v$ is represented by the logarithm of its absolute value and its sign. Thus,
\begin{subequations}\label{eq:2_1}
	\begin{align}
		v \longleftrightarrow \underline{V} & = \left(V, s_v \right) \subtag{a} \label{eq:2_1_a}\\
		V & = \log_2\left(\left| v\right|\right) \subtag{b} \label{eq:2_1_b}\\
		s_v & = \mathrm{sign}(v) \subtag{c} \label{eq:2_1_c}
	\end{align}
\end{subequations}
where $\mathrm{sign}(v) = 1$ if $v>0$ and 0 otherwise. Note that the radix of the logarithm does not change the important properties of LNS, but using radix $2$ leads to bit-shift approximations as described in Section \ref{sec:appr}.
Multiplication in the linear-domain becomes addition in log-domain
\begin{subequations} \label{eq:2_2}
	\begin{align}
		u = xy \longleftrightarrow \underline{U} & = \underline{X} \boxdot \underline{Y} \subtag{a} \label{eq:2_2_a}\\
		U & = X + Y \subtag{b} \label{eq:2_2_b}\\
		s_u & = \overline{(s_x \veebar s_y)} \subtag{c} \label{eq:2_2_c}
	\end{align}
\end{subequations}
where $\veebar$ denotes the exclusive OR operation and $\overline{s}$ denotes the compliment of the binary variable $s$.  
We define addition in log-domain as follows
\begin{subequations} \label{eq:2_3}
	\begin{align}
		z &= x+y  \longleftrightarrow \underline{Z}  = \underline{X} \boxplus \underline{Y} \subtag{a} \label{eq:2_3_a}\\
		Z & = \begin{cases}
     	\max(X,Y) + \Delta_{+}\left(\left|X-Y\right|\right)   & s_x = s_y\\
     	\max(X,Y) + \Delta_-\left(\left|X-Y\right|\right)     & s_x \neq  s_y
     	\end{cases} \subtag{b} \label{eq:2_3_b}\\
     	s_z & = \begin{cases} s_x & X>Y\\ s_y & X \leq Y \end{cases} \subtag{c} \label{eq:2_3_c}
	\end{align}
\end{subequations}
where the $\Delta$ terms exact representation of addition in log-domain in the extended real numbers $R^+$. The functional expression of $\Delta$ terms are
\begin{subequations} \label{eq:2_4}
	\begin{align}
		\Delta_{+}(d) &= \log_2\left(1 + 2^{-d}\right) && d\geq 0 \subtag{a} \label{eq:2_4_a}\\
		\Delta_{-}(d) &= \log_2\left(1 - 2^{-d}\right) && d\geq 0 \subtag{b} \label{eq:2_4_b}
	\end{align}
\end{subequations}
To reduce the computational complexity of calculating $\Delta$, we describe two different approximations in Section \ref{sec:appr} that induce approximate addition in the log-domain. One can extend the above concepts to define log-domain subtraction as
\begin{equation} \label{eq:2_5}
	t = x-y  \longleftrightarrow \underline{T}  = \underline{X} \boxminus \underline{Y} = \underline{X} \boxplus \left(Y, \overline{s_y}\right)
\end{equation}
Exponentiation translates to a multiplication in the log-domain. This operation is non-commutative as in general $a^b \neq b^a$. We define log-domain exponentiation when exponentiating on a positive radix $x > 0$,
\begin{equation} \label{eq:2_6}
	w = x^y  \longleftrightarrow \underline{W}  = \left( y  X, 1\right)
\end{equation}

\section{Approximate Log-Domain Addition}
\label{sec:appr}
\begin{figure}
    \centering
    \includegraphics[width = 0.48\textwidth]{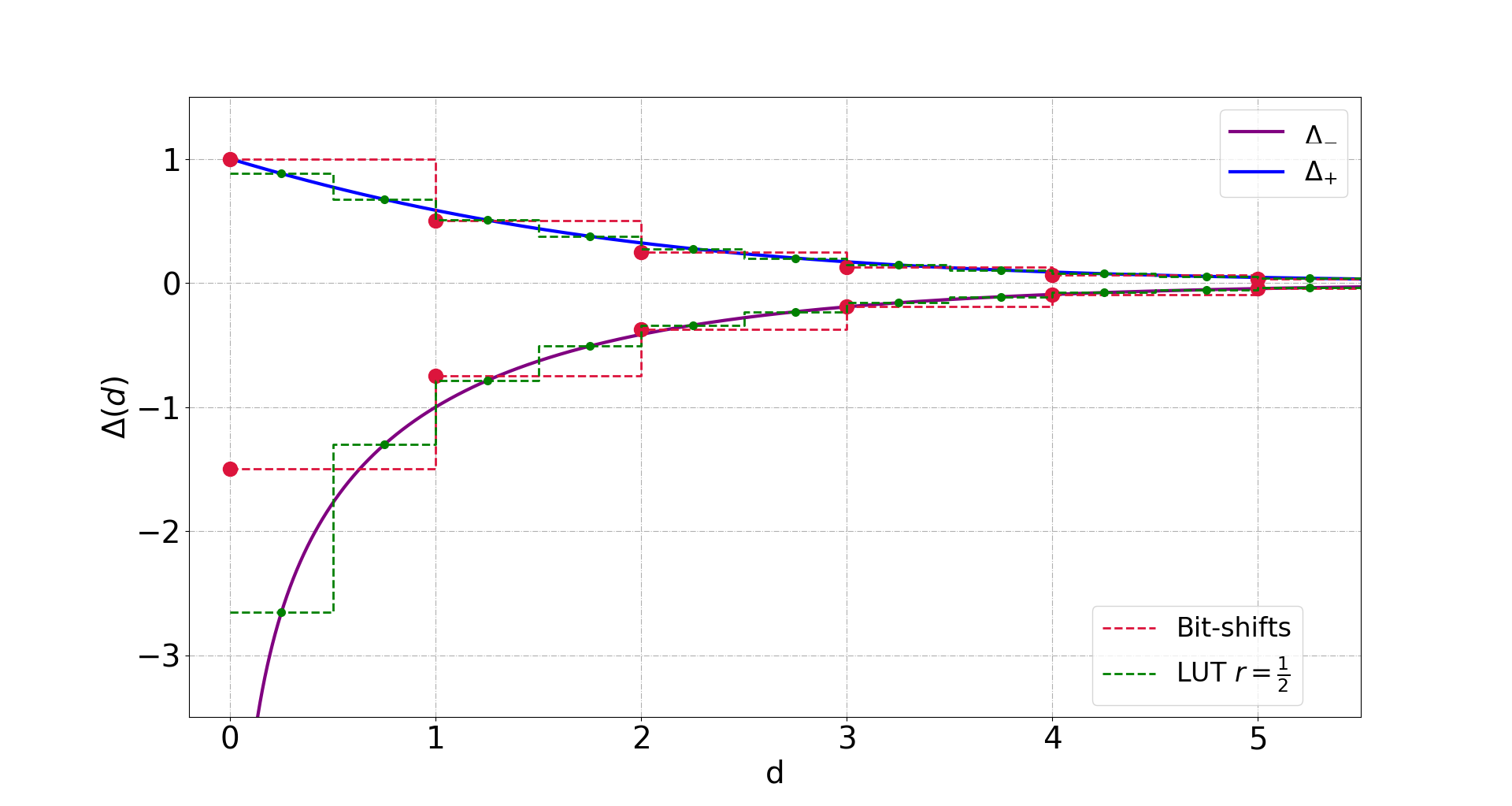}\\
    \caption{Approximation to $\Delta_\pm(d)$ (table size is 20).}
    \label{fig:delta_curves}
\end{figure}
It is clear from (\ref{eq:2_2}) that LNS processing reduces the complexity of multiplication, but 
the $\Delta$ terms in (\ref{eq:2_3}) associated with log-domain addition are much more complex to implement than standard addition. Motivated by the fact that the training process is inherently noisy (e.g., gradient noise, finite precision effects, etc.) we propose low-complexity approximations of the $\Delta$ terms in (\ref{eq:2_3}). Specifically we seek the simplest such approximations that do not significantly degrade the overall accuracy of the trained networks. Look-up tables provide a natural approach to approximating the $\Delta$ terms. In this paper we consider simple LUTs for $\Delta_{\pm}(d)$ wherein the \textit{dynamic range} of $d$ supported is $\left[ 0, d_{\mathrm{max}}\right]$ and the resolution is $r$. Specifically, each unit interval within the dynamic range has $1/r$ points uniformly sampled from $\Delta_{\pm}(d)$. This concept is shown in Fig. \ref{fig:delta_curves}. Note that the size of the LUT is $d_{\mathrm{max}} / r$.  

A bit-shift approximation for $\Delta_+(d)$ was suggested in \cite{stanford_arxiv}. This can be generalized using 
\begin{align}
	    \log_e\left(1 \pm x\right)& \approx \pm  x &&~0 \leq x \ll 1 \label{eq:3_1_a}
\end{align}
which, together with (\ref{eq:2_4}) implies that
\begin{equation}
\Delta_\pm(d) \approx \pm 2^{-d} \left(1+2^{-1}-2^{-4} + \cdots \right)  \label{eq:3_2_b}
\end{equation}
where the term in the parenthesis is the fixed-point approximation of $\log_2(e)$. 
This can be viewed as a bit-shift approximation as all operations involve multiplication by powers of two. In particular, the most accurate approximations of this form are
\begin{subequations}\label{delta_bs}
    \begin{align}
        \Delta_+(d) &\approx\BS{1}{-d} \label{delta_plus_bs}\\
        \Delta_-(d) &\approx-\BS{1.5}{-d} \label{delta_minus_bs}
    \end{align}
\end{subequations}
 where $\BS{a}{b} =a 2^b$ corresponds to a bit-shift in the binary representation of $a$ by $b$ positions to the left. The bit-shift approximations in (\ref{delta_bs}) are shown in Fig. \ref{fig:delta_curves}. Finally, note that these bit-shift approximations are equivalent to a LUT with $r=1$ and dynamic range set by the maximum value of $d$ possible with the bit-width of the fixed-point representation.
 

\section{Log-Domain DNN Training}
\label{sec:lognn}
Much of the computation associated with the feedforward and backpropagation operations are based on matrix multiplication. These can be implemented directly using the operations in 
Sections \ref{sec:lns}-\ref{sec:appr} 
\begin{equation}\label{eq:4_2_1}
    z_i=\sum_j w_{i,j} x_{j} + b_i\longleftrightarrow Z_i=\mathop{\mathlarger{\mathlarger{\mathlarger{\boxplus}}}}\limits_{j} \underline{W}_{i,j}\boxdot\underline{X}_j\boxplus\underline{B}_i
\end{equation}
In this section we describe log-domain versions of the other significant operations in the training of a DNN. While the general approach is applicable to all types of neural networks, we focus on multi-layer perceptrons (MLPs) to demonstrate the concept. Specifically, these are: (i) activation functions, (ii) weight initialization, (iii) soft-max operations, and (iv) dataset conversion. We also briefly discuss issues with fixed-point representation.  

\textbf{Activation Functions:} It is most efficient to determine the log-domain equivalent of the desired activation function and implement that directly in the log domain processing. In the numerical results that follow, a leaky-ReLU \cite{prelu} activation is used. This translates to a 
\textit{log-leaky ReLU} (llReLU) in the log-domain
\begin{equation}\label{eq:4_1_1}
    g_{\mathrm{llReLU}}\left(\left(X, s_x\right)\big|\beta\right)=\begin{cases} \left(X, s_x\right) & s_x=1 \\ \left(X+\beta, s_x\right) & s_x=0 \end{cases}
\end{equation}
where $\beta$ is a single hyper-parameter associated with this activation function. In back-propagation, the derivative of the activation function is required and, in this case of leaky-ReLU, the derivative is simple to implement directly in the log-domain.  

\textbf{Weight Initialization:} Weights are conventionally initialized according to some specified distribution and it is most efficient to translate this distribution to the log-domain and initialize the log-domain weights accordingly. The probability density function used for weight initialization is typically symmetric around zero -- i.e., $f_w\left( x\right) = f_w\left( -x\right)$. We consider such symmetric distributions for which the sign in log domain is Bernoulli distributed, equally likely to be 0 or 1. The distribution for $W=\log_2\left|w\right|$, can be determined using standard change of measure approaches from probability as
\begin{equation}
f_W\left(y\right) =  2^{y+1} \log_e(2)  f_w\left(2^y\right)
\end{equation}

\textbf{Soft-max Layer:} In classification tasks, it is common to use a final soft-max layer with a cross-entropy cost \cite{deep_learning_book}. 
The soft-max and the associated gradient initialization are
\begin{subequations}\label{eq:4_2_2}
    \begin{align}
        p_{ij} &= \frac{e^{a_{ij}}}{\sum_{j=1}^{N} e^{a_{ij}}}\subtag{a}\label{eq:4_2_2_a}\\
        \delta_{ij} &= p_{ij} - y_{ij}\subtag{b}\label{eq:4_2_2_b}
    \end{align}
\end{subequations}
where $y_{ij}$ is the one-hot encoded label. In the log-domain, this corresponds to 
\begin{subequations}\label{eq:4_2_3}
    \begin{align}
    \log_2p_{ij} &= \left( a_{ij}\log_2e\right) - \mathop{\mathlarger{\mathlarger{\mathlarger{\boxplus}}}}\limits_{j=1}^{N}\left(a_{ij}\log_2e, 1\right)\label{eq:4_2_3_b} \\
        \left(\log_2\left|\delta_{ij}\right|, s_{\delta_{ij}}\right) &= \underline{P}_{ij} \boxminus \left(\log_2\left| y_{ij}\right|, s_{y_{ij}}\right) \label{eq:4_2_3_a}
    \end{align}
\end{subequations}

\textbf{Dataset Conversion:} The dataset used for training or the inputs used during inference also need to be converted to the log-domain. In the numerical results that follow, this was done with off-line pre-processing using floating point operations. In a real-time application this conversion requires computing $\log_2\left( \sum_i 2^i \right)$ and therefore could also be performed using the (approximate) operations in Sections \ref{sec:lns}-\ref{sec:appr}.

\textbf{Fixed-Point Implementation:} A fixed-point representation of $x$ in the LNS described in Section \ref{sec:lns} with $q_i$ integer bits and $q_f$ faction bits will have a total of $W_{\mathrm{log}} = 2 + q_i +q_f$ bits owing to the single bit to represent $s_x$ and the bit for the sign of $X$. In analysis omitted for brevity, we show that the number of bits required in the log-domain to ensure the same range and precision as a given fixed-point representation in the linear-domain is 
\begin{equation}
        W_{\mathrm{log}} \geq 1 + \max \big(\ceil{\log_2 \big(b_i+ 1\big)}, \ceil{\log_2 b_f}\big) + W_{\mathrm{lin}} \label{log_bw}
\end{equation}
where $W_{\mathrm{lin}}$ is the bit-width in the linear domain, comprised of $1$ sign bit along with $b_i$ and $b_f$ integer and fractional bits, respectively. For a typical value of 16-bit precision, with $b_i=4$ and $b_f=11$, $W_{\mathrm{log}}=21$ is required to guarantee the same precision and dynamic range. The analysis leading to (\ref{log_bw}) is worst-case and our numerical experiments, summarized in Section \ref{sec:results}, suggest that $W_{\mathrm{log}} \approx W_{\mathrm{lin}}$ suffices in practice.

\begin{figure}[!t]
    \centering
    \subfloat[MNIST\label{subfig:1a}]{\includegraphics[width=4.2cm]{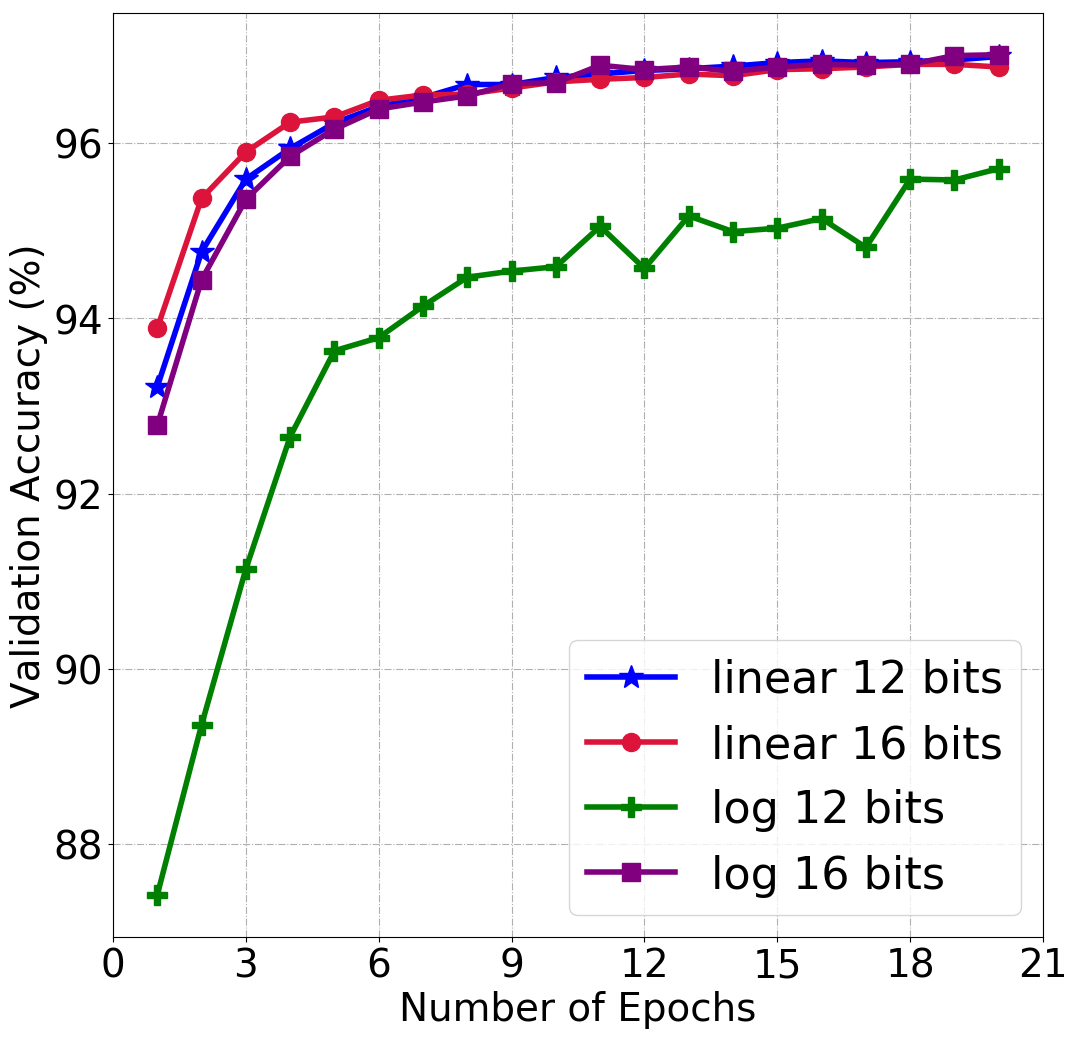}}
    \subfloat[Fashion-MNIST\label{subfig:1b}]{\includegraphics[width=4.2cm]{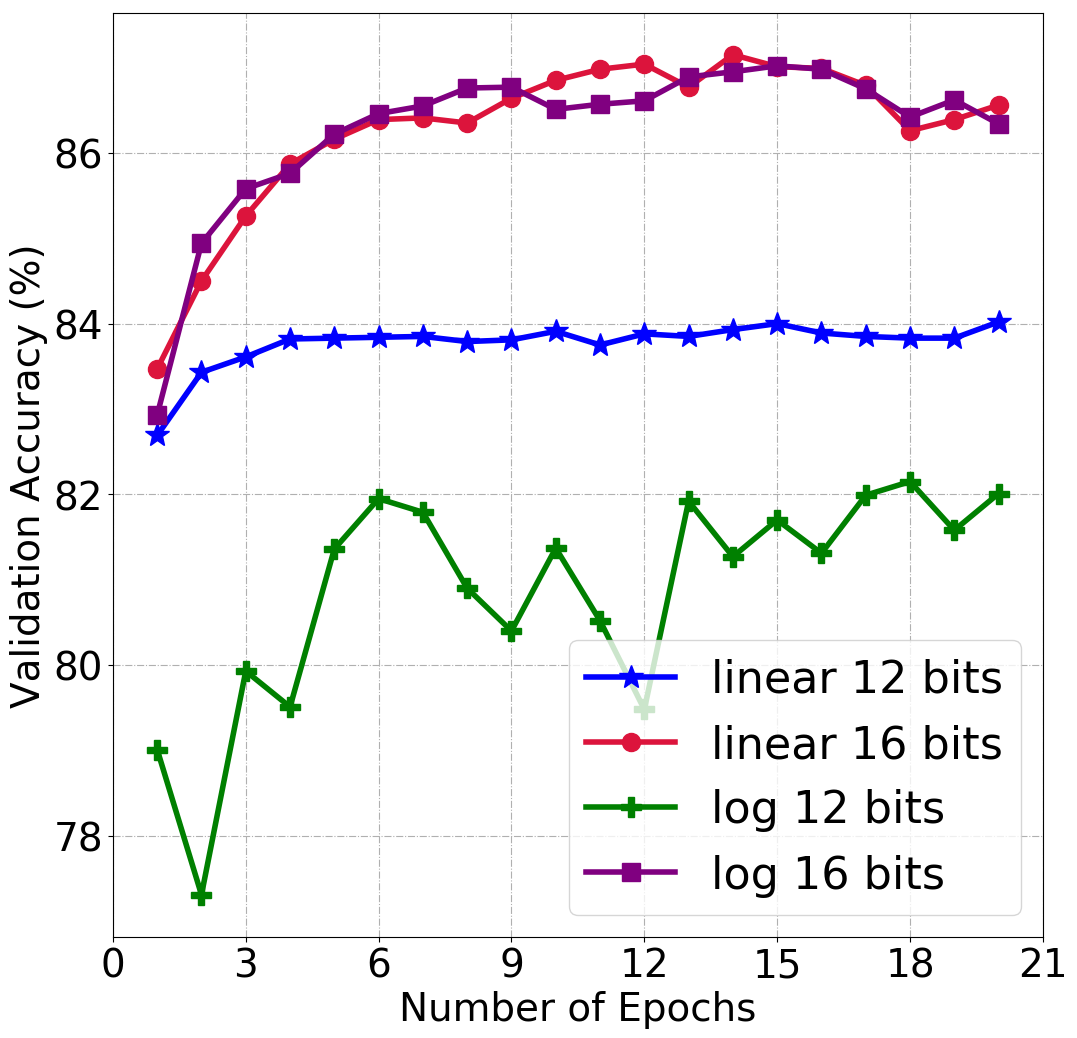}}\vfill
    \subfloat[EMNIST-digits\label{subfig:1c}]{\includegraphics[width=4.2cm]{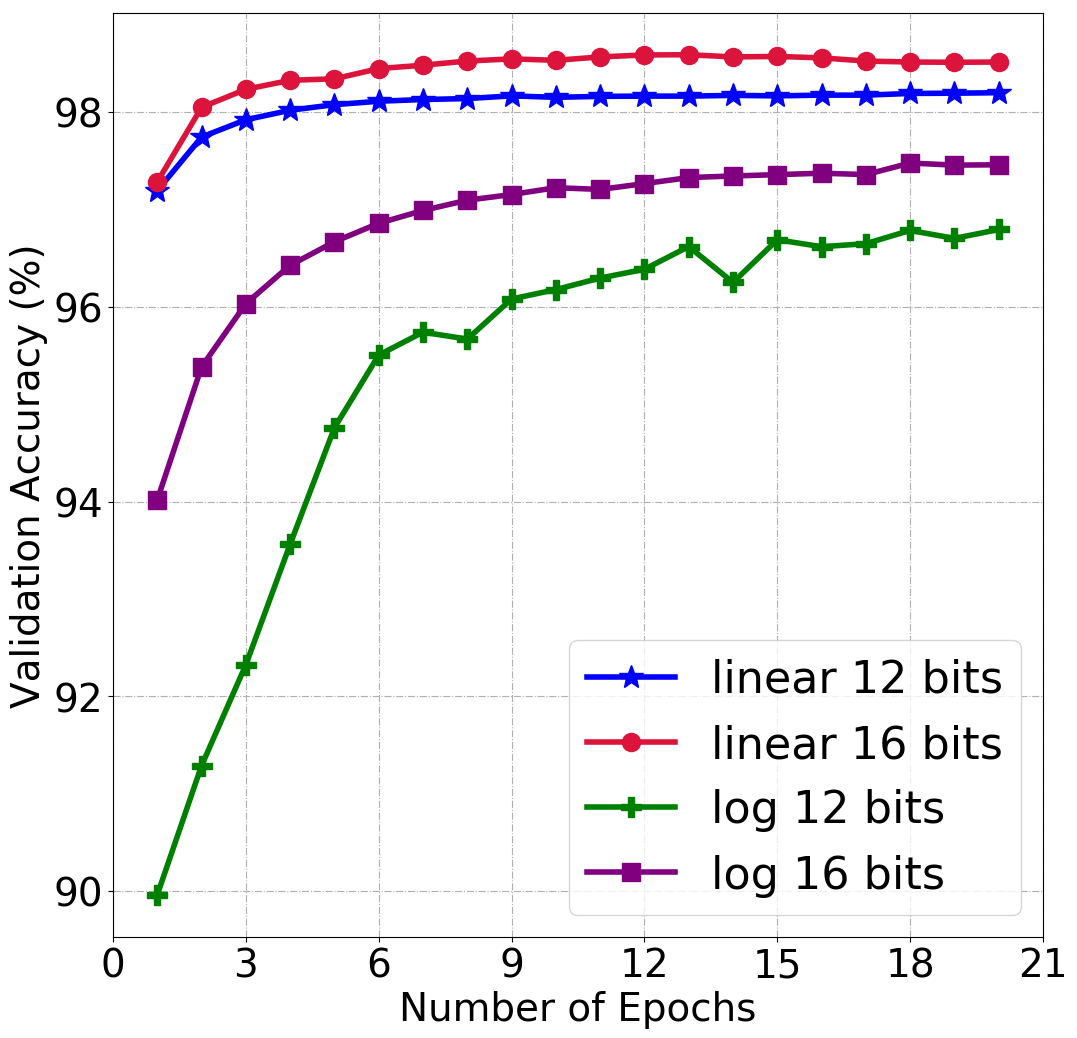}}
    \subfloat[EMNIST-letters\label{subfig:1d}]{\includegraphics[width=4.2cm]{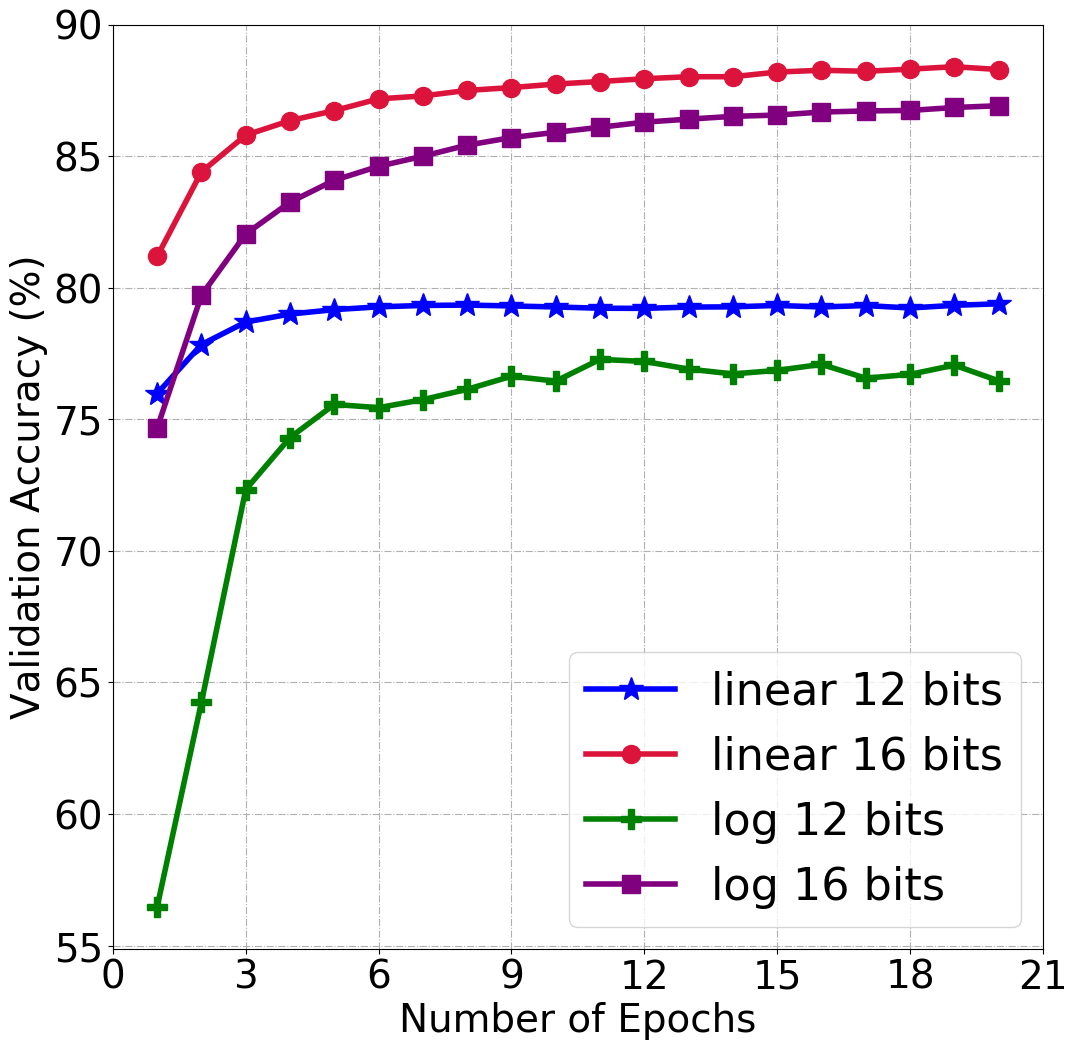}}
    \caption{Validation accuracy learning curves comparing 12 and 16 bit log-domain training with 12 and 16-bit linear training. Log-training was performed using a 20 element table ($d_{\mathrm{max}}=10$, $r=1/2$) for all operations except the soft-max which employed a 640 element table ($d_{\mathrm{max}}=10$, $r=1/64$). } 
    \label{fig:validation_accuracy}    
\end{figure}



\section{Numerical Experiments}
\label{sec:results}



The neural network trained is an MLP with one input layer of $784$ neurons, one hidden layer of $100$ neurons, and one soft-max layer with number of neurons equal to the number of classes for the given dataset. Stochastic gradient descent was used with mini-batch size of $5$ and learning rate of $0.01$. The weight decay regularization constant was optimized for each individual dataset. In general, $12$-bit implementation needed a larger regularization constant than the $16$-bit implementations. The activation function used in the hidden layer for the linear baselines is leaky-ReLU \cite{prelu} and llReLU for the log experiments. When approximating $\Delta_-$, its value at $0$ is set to be the most negative number the fixed point setting can represent. Our experimental results are available online at \cite{our_code}, where the log-domain core has been implemented in C with a Python callable wrapper.

Four balanced datasets were used for experiments: MNIST \cite{mnist}, Fashion-MNIST \cite{fashion_mnist} (FMNIST), EMNIST-Digits (EMNISTD), and EMNIST-Letters (EMNISTL) \cite{emnist}. All datasets contain $8$-bit encoded, gray-scale square images with $784$ pixels. 
MNIST and FMNIST each have $10$ output classes and comprise $6,000$ training images and $10,00$ test images per class. 
EMNISTD has $24,000$ and $4,000$ images per each of its $10$ classes for training and test, respectively. 
EMNISTL has $4,800$ and $800$ images per each of its $26$ classes for training and test, respectively. Validation data was held back from the training datasets with a 1:5 ratio.  

Learning curves are shown in Figure \ref{fig:validation_accuracy} for finite precision linear and log-domain training. In the linear-domain, using 16-bit fixed-point representation, with $11$ bits for the fractional part, was found to provide negligible degradation relative to floating point. In the log-domain, 16-bit representations use $10$ fractional bits (i.e., owing to the extra bit needed for the sign). When performing experiments on $12$ bit systems, the number of fractional bits is kept at $7$ and $6$ for linear and log-domain, respectively. For the LUT-based approximations in the log-domain, we first minimized the table sizes need empirically. First, high-resolution was used and the minimum value of dynamic range required for good overall accuracy was determined to be $d_{\mathrm{max}} = 10$. Next, fixing the dynamic range to 10, we varied the resolution and determined that $r=1/2$ was required to achieve minimal degradation relative to linear-domain results. We found that the log-domain implementation of the soft-max was more sensitive to approximation errors and the results in Figure \ref{fig:validation_accuracy} utilize a resolution of $r=1/64$ for the soft-max processing. Table \ref{tab:bitshift_results} provides a comparison of test-set accuracy for fixed-point linear processing and full log-domain training and inference with various approximations.  
\begin{table}[t!]
    \centering
    \begin{minipage}{\columnwidth}
    \resizebox{\columnwidth}{!}{
    \begin{tabular}{||c||c||c|c||c|c||c|c||}
        \hhline{--------}
        \multirow{4}{*}{\textbf{Datasets}} & \multirow{4}{*}{Float} & \multicolumn{2}{c||}{\textbf{Linear-domain}} & \multicolumn{2}{c||}{\textbf{Log-domain}} & \multicolumn{2}{c||}{\textbf{Log-domain}} \\
         & & \multicolumn{2}{c||}{fixed-point} & \multicolumn{2}{c||}{fixed-point} & \multicolumn{2}{c||}{fixed-point}\\
         & & \multicolumn{2}{c||}{} & \multicolumn{2}{c||}{look-up tables} & \multicolumn{2}{c||}{bit-shifts}\\
        \cline{3-8}
         & & $12$b & $16$b & $12$b & $16$b & $12$b & $16$b\\
        \hhline{========}
        MNIST & 97.4 & 97.3 & 96.9 & 96.0 & 97.2 & 95.5 & 96.5 \\[0.5ex]
        \hline
        FMNIST & 87.1 & 82.8 & 88.0 & 80.5 & 87.1 & 79.3 & 85.7 \\[0.5ex]
        \hline
        EMNISTD & 98.6 & 98.3 & 98.7 & 96.9 & 97.5 & 96.2 & 97.4 \\[0.5ex]
        \hline
        EMNISTL & 88.1 & 79.7 & 88.7 & 76.4 & 86.7 & 73.7 & 82.5 \\[0.5ex]
        \hline
    \end{tabular}
    }
    \end{minipage}
    \caption{Test set accuracy (\%) at 20 epochs. Fixed-point log-domain results are for LUT and bit-shift approximations to log-domain adds (i.e., $\Delta_{\pm}(\cdot)$ approximations).}
    \label{tab:bitshift_results}
\end{table}

\section{Conclusions}
\label{sec:conc}
 Our results demonstrate that all training and inference processing associated with a neural network can be performed using logarithmic number system with approximate log-domain additions, thus allowing a hardware implementation without multipliers. In particular, approximating the log-domain addition using a $\max(\cdot)$, add, and an approximation to the $\Delta$-term based on a LUT yields only modest degradation in classification accuracy as compared to that of linear processing. Similar to linear processing, we conclude that 16-bit fixed-point representations are sufficient to approach the classification accuracy associated with floating point computations. We also found that a LUT of size 20 was sufficient and that a simple bit-shift approximation, which can be viewed as equivalent to a smaller table, also provides good classification performance in many cases.

Future areas of research include application to larger convolutional neural networks popular in computer vision and co-optimization of $\Delta$-term approximations considering classification accuracy and hardware complexity. While this work demonstrates the potential of this approach, the circuit implementation complexity of the approximate log-domain adder must be significantly lower than that of a multiplier in order for the approach to be desirable in practice.

\bibliographystyle{IEEEbib}
\bibliography{main}

\end{document}